\documentclass[runningheads]{llncs}
\usepackage{graphicx}
\newcommand\correspondingauthor{\thanks{Corresponding author.}}

\usepackage{balance}
\usepackage{cite}
\usepackage{ amssymb }
\usepackage{amsmath}
\usepackage[margin=1.1in]{geometry}   
\usepackage{makecell}
\usepackage{algorithm}
\usepackage[noend]{algorithmic}

\newtheorem{mydef}{Definition}
\usepackage{subfig}
\usepackage{enumerate}
\usepackage{multirow,tabularx}
\usepackage{xcolor}
\newcommand*\samethanks[1][\value{footnote}]{\footnotemark[#1]}

\begin{document}
\title{Continual Learning with Differential Privacy}
%
%
\author{Pradnya Desai\inst{1}\thanks{These two authors contributed equally.} \and
Phung Lai\inst{1}\samethanks  \and
NhatHai Phan\inst{1}\correspondingauthor \and My T. Thai\inst{2} }
\authorrunning{P. Desai et al.}
%
\institute{New Jersey Institute of Technology, USA \\ 
 \email{$\{$pnd26,tl353,phan$\}$@njit.edu}
\and
University of Florida, USA \\
\email{mythai@cise.ufl.edu}}
\maketitle              
\begin{abstract}
In this paper, we focus on preserving differential privacy (DP) in continual  learning (CL), in which we train ML models to learn a sequence of new tasks while memorizing previous tasks. We first introduce a notion of continual adjacent databases to bound the sensitivity of any data record participating in the training process of CL. Based upon that, we develop a new DP-preserving algorithm for CL with a data sampling strategy to quantify the privacy risk of training data in the well-known Averaged Gradient Episodic Memory (A-GEM) approach by applying a moments accountant. Our algorithm provides formal guarantees of privacy for data records across tasks in CL. Preliminary theoretical analysis and evaluations show that our mechanism tightens the privacy loss while maintaining a promising model utility.

\keywords{Continual learning   \and Differential privacy \and Deep learning.}
\end{abstract}
\section{Introduction}

The ability to acquire new knowledge over time while retaining previously learned experiences, referred to as continual learning (CL), brings machine learning closer to human learning \cite{ratcliff1990connectionist,schwarz2018progress,aljundi2019task}. More specifically, given a stream of tasks, CL focuses on training a machine learning (ML) model to quickly learn a new task by leveraging the acquired knowledge after learning previous tasks under a limited amount of computation and memory resources \cite{lopez2017gradient,rusu2016progressive}. As a result, the main challenge of existing CL algorithms is that they can quickly be suffered by catastrophic forgetting. 

Also, memorizing previous tasks while learning new tasks further exposes CL models to adversarial attacks, especially model and data inference \cite{shokri2017membership,fredrikson2015model,wang2015regression}. CL models can disclose private and susceptible information in the training set, such as healthcare data \cite{zia2020application,alnemari2017adaptive,kartal2019differential}, financial records \cite{wu2019value,zhu2017differential}, and bio-medical images \cite{plis2014deep,helmstaedter2013connectomic}. Continuously accessing the data from the previously learned tasks, either stored in episodic memories \cite{chaudhry2018efficient,riemer2018learning,abati2020conditional,tao2020few,rajasegaran2020itaml} or produced from generative memories \cite{shin2017continual,wu2018memory,ostapenko2019learning}, incurs additional privacy risk compared to a single ML model trained on a single task.
However, there is still a lack of scientific study to protect private training data in CL algorithms.



Motivated by this, we propose to preserve differential privacy (DP) \cite{dwork2006calibrating}, offering rigorous privacy protection as probabilistic terms for the training data in CL. Merely employing existing DP-preserving mechanisms can either cause a significantly large privacy loss or quickly exhaust the limited computation and memory resources in learning new tasks while memorizing previous tasks through either episodic or generative memories. Thus, effectively and efficiently preserving DP in CL remains a mostly open problem.


\textbf{Key contributions}. To effectively bound the DP privacy loss in CL, we first define continual adjacent databases (\textbf{Def. \ref{CAD}}) to capture the impact of the current task's data and the episodic memory on the privacy loss and model utility. Based upon that, we incorporate  a moments accountant \cite{abadi2016deep} into  the Averaged Gradient Episodic
Memory (A-GEM) algorithm \cite{chaudhry2018efficient} in a new \textbf{DP-CL} algorithm to preserve DP in CL.

Our idea is to configure the episodic memory in A-GEM as independent mini-memory blocks. We store a subset of training data of the current task in a mini-memory block with an associated task index in the episodic memory for each task. At each training step, we compute reference gradients on the mini-memory blocks independently. The reference gradients will be used to optimize the process of memorizing previously learned tasks as in A-GEM. More importantly, by keep tracking of the task and mini-memory block index, we can leverage a moments accountant to estimate the privacy cost spent on each mini-memory block. Based upon this, we derive a new strategy (\textbf{Lemma \ref{lmrandom}}) to bound DP loss in the whole CL process while maintaining the computation efficiency of the A-GEM algorithm. 

To our knowledge, our proposed mechanism establishes the first formal connection between DP and CL. Experiments conducted on the permuted MNIST dataset \cite{goodfellow2013empirical} and the Split CIFAR \cite{zenke2017continual} show promising results in preserving DP in CL, compared with baseline approaches. 



\section{Background} 
In this section, we revisit continual learning, differential privacy, and introduce our problem statement. The goal of CL is to learn a model through a sequence of tasks $T =[t_i]_{i \in [1, N]}$ such that the learning of each new task will not cause forgetting of the previously learned tasks. Let $\mathcal{D}_{\mathcal{T}}$ be the dataset at task $\mathcal{T}$ consisting of $S_{\mathcal{T}}$ samples, each of which is a sample $x \in \mathbb{R}^d$ associated with a label $y$.
Each $y$ is a one-hot vector of $C$ categories: $y = [ y_c ]_{c \in [1, C]}$. A classifier outputs class scores $f: \mathbb{R}^d \rightarrow \mathbb{R}^C$ mapping an input $x$ to a vector of scores $f(x) = [f_c(x)]_{c \in [1, C]}$ s.t. $\forall c \in [1,C]: f_c(x) \in [0,1]$ and $\sum_{c=1}^C f_c(x) = 1$. The class with the highest score is selected as the predicted label for the sample. The classifier $f$ is trained by minimizing a loss function $\mathcal{L}(f(x), y)$ that penalizes mismatching between the prediction $f(x)$ and the original value $y$.

\textbf{Averaged  Gradient Episodic Memory (A-GEM) \cite{chaudhry2018efficient}}.  There is a sequence of tasks $[ t_i]_{i \in [1, \mathcal{T} - 1]}$ that have been learnt, where $\mathcal{T} < N$. The goal is to train the model at the current task ${\mathcal{T}}$ so that it minimizes the loss on the task  ${\mathcal{T}}$ and does not forget previous learned tasks $i < \mathcal{T}$. 
The key feature of A-GEM is to store a subset of data from task $i$, denoted as $\mathcal{M}_i$, in an episodic memory $\mathcal{M}$. 
Then the algorithm ensures that the loss on an average episodic memory across all the previously learned tasks, i.e., $\mathcal{M} = \cup_{i < \mathcal{T}} \mathcal{M}_i$, does not increase at every training step. In A-GEM, the objective function of learning the current task ${\mathcal{T}}$ is: 
\begin{align}
& \theta^{\mathcal{T}} = \min_{\theta} \mathcal{L}\big(f(\theta^{\mathcal{T}-1}, \mathcal{D}_{\mathcal{T}})\big) \nonumber \\ 
& \text{ s.t. } \mathcal{L}\big(f(\theta^{\mathcal{T}}, \mathcal{M})\big) \le \mathcal{L} \big(f({\theta}^{\mathcal{T}-1}, \mathcal{M})\big) \text{ for } \mathcal{M} = \cup_{i < \mathcal{T}} \mathcal{M}_i \label{agem} 
\end{align} 
where $\theta^{\mathcal{T} - 1}$ is the values of model parameters $\theta$ learned after training the task $\mathcal{T}-1$, and $\mathcal{L}\big(f(\theta^{\mathcal{T}-1}, \mathcal{D}_{\mathcal{T}})\big) = \frac{1}{|\mathcal{D}_{\mathcal{T}}|}{\sum_{x \in \mathcal{D}_{\mathcal{T}}}\mathcal{L}\big(f(\theta^{\mathcal{T}-1}, x)\big)}$.

The constrained optimization problem of Eq.\ref{agem} can be approximated quickly and the updated gradient $\tilde{g}$ is as follows:
\begin{equation}\label{gradient}
 \tilde{g} = g - \frac{g^Tg_{ref}}{ g^T_{ref} g_{ref} } g_{ref}
\end{equation}
where $g$ is the proposed gradient update on $\mathcal{T}$ and $g_{ref}$ is the reference gradient computed from the episodic memory $\mathcal{M}$ from previous tasks. 

\textbf{Differential Privacy  \cite{dwork2014algorithmic,dwork2008differential,mcsherry2007mechanism,xu2019ganobfuscator}}.
To avoid the training data leakage, DP guarantees to restrict what the adversaries can learn from the training data given the model parameters by ensuring similar model outcomes
with and without any single data sample in the dataset. The definition of DP is as follows:

\begin{mydef}{$(\epsilon, \delta)$-DP \cite{dwork2006calibrating}.} A randomized algorithm $A$ fulfills $(\epsilon, \delta)$-DP, if for any two adjacent databases $D$ and $D'$ differ at most one sample, and for all outcomes $\mathcal{O} \subseteq Range(A)$, we have: 
\begin{equation}
Pr[A(D) = \mathcal{O}] \leq e^{\epsilon} Pr[A(D') = \mathcal{O}] + \delta 
\end{equation}
where $\epsilon$ is the privacy budget and and $\delta$ is the broken probability. 
\label{Different Privacy} 
\end{mydef}

The privacy budget $\epsilon$ controls the amount by which the distributions induced by $D$ and $D'$ may differ. A smaller $\epsilon$ enforces a stronger privacy guarantee. 
The broken probability $\delta$ represents the improbable ``bad" events in which an adversary can infer whether a particular data sample belongs to the training data, which is possible when the probability $\le \delta$. 



\textbf{DP in Continual Learning}. There are several works of DP in CL \cite{farquhar2019differentially,phan2019private}.  In \cite{farquhar2019differentially}, the authors train a DP-GAN \cite{zhang2018differentially} to approximate the distribution of the past datasets. They leverage a small portion of public data (i.e., the data that does not need to keep private) to initialize and train the GAN in the first few  iterations of each task, then  continue training the GAN model under DP constraint. The trained generator produces adversarial examples imitating real examples of past tasks. Then, the adversarial examples are employed to supplement the actual data of the current training task. DPL2M \cite{phan2019private} perturbs the objective functions using a DPAL mechanism \cite{phan2020scalable,DBLP:journals/corr/abs-1903-09822} and applies A-GEM to optimize the perturbed objective function.  However, there is a lack of a concrete definition of adjacent databases with unclear or not well-justified DP protection in \cite{farquhar2019differentially,phan2019private}. Different from existing works, we provide a formal DP protection for CL models.



\section{Continual Learning with DP}

 \begin{figure}[t]
      \centering
      \includegraphics[scale=0.5]{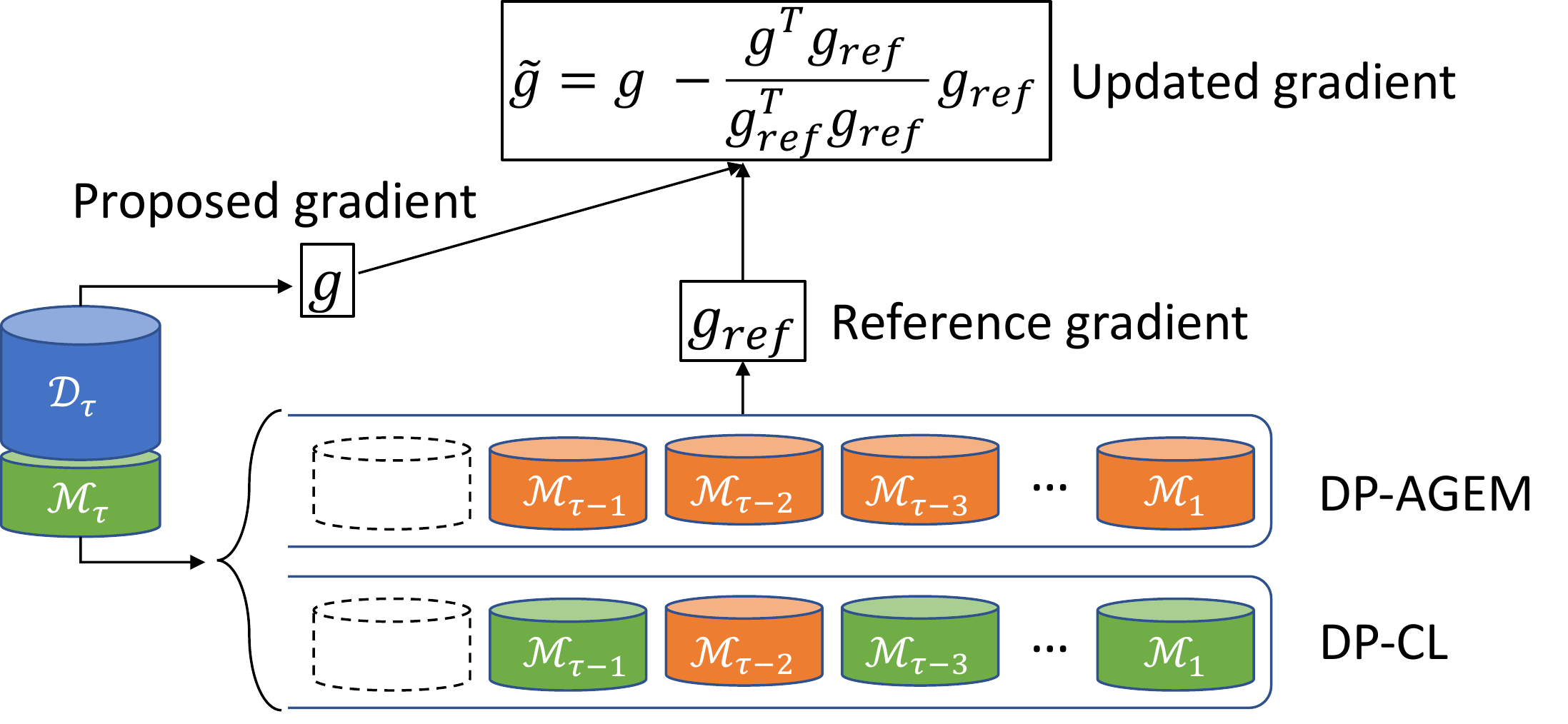} 
      \caption{ DP in CL protects privacy for a stream of different tasks. The updated gradient $\tilde{g}$ is computed based on 1) $g_{ref}$ computed from the episodic memory $\mathcal{M}$ for previous tasks and 2) $g$ of the current task $\mathcal{T}$ using  our proposed algorithm DP-CL. Here, blue box indicates training data of task $\mathcal{T}$, orange boxes and green boxes indicate mini-memory blocks in $\mathcal{M}$ in which the orange ones are used to compute $g_{ref}$. } 
      \label{dp-agem}
 \end{figure}
 
This section establishes a connection between differential privacy and continual learning. We first propose a definition of continual adjacent databases in CL, as follows: Two databases $D$ and $D'$ are continual adjacent if they differ in a single sample of the training data and differ in a single sample of the episodic memory across all the tasks. The definition is presented as follows:
  \begin{mydef}{Continual Adjacent Databases.} Two databases $D = (\mathcal{D}, \mathcal{M})$ and $D'= (\mathcal{D}', \mathcal{M}')$, where $\mathcal{D} = \cup_{i=1}^N \mathcal{D}_i $, $\mathcal{D}' =\cup_{i=1}^N \mathcal{D}'_i$, $\mathcal{M} = \cup_{i=1}^N \mathcal{M}_i $, and $\mathcal{M}' = \cup_{i=1}^N \mathcal{M}'_i$, are called continual adjacent if: $\|\mathcal{D}-\mathcal{D}'\|_1 \le 1$ and $\|\mathcal{M}-\mathcal{M}'\|_1  \le 1$.
  \label{CAD}
 \end{mydef}


\textbf{A Naive Algorithm}. Based upon Definition \ref{CAD}, a straightforward approach, called DP-AGEM, is to simply apply a moments accountant \cite{abadi2016deep} into A-GEM \cite{chaudhry2018efficient}, to preserve DP in CL. At each task $\mathcal{T}$, we divide the dataset $D_{\mathcal{T}}$  into  $D_{\mathcal{T}}^{train}$ and $D_{\mathcal{T}}^{ref}$ such that $D_{\mathcal{T}}^{train}$ and $D_{\mathcal{T}}^{ref}$ are disjoint: $D_{\mathcal{T}}^{train} \cap D_{\mathcal{T}}^{ref} = \emptyset$. By using the training data $\mathcal{D}^{train}_\mathcal{T}$ with a sampling rate $p$, DP-AGEM computes a proposed gradient $g$, which is bounded  by a predefined $l_2$-norm clipping bound $\beta$. It is beneficial in real-world to keep track of the privacy budget spent on each task independently, and the total privacy budget used in the entire training process. To achieve this, in computing the reference gradients $g_{ref}$, the algorithm first randomly samples data from all the data samples in the episodic memory $\mathcal{M}$ with a sampling probability $q$. Given a particular $D_{i}^{ref}$ ($i \in [1, \mathcal{T} - 1]$) in the episodic memory, the sampled data is used to compute a reference gradient $g^{i}_{ref}$, which is clipped with the $l_2$-norm bound $\beta$. Then Gaussian mechanism is employed to inject random Gaussian noise $\mathcal{N}(0, \sigma^{2}\beta^{2}I)$ with a predefined hyper-parameter $\sigma$ into both $g$ and $g^{i}_{ref}$. The reference gradient $g_{ref}$ is the average of all the reference gradients computed on each $D_{i}^{ref}$, as follows: $g_{ref} = \frac{1}{\mathcal{T} - 1} \sum_{i \in [1, \mathcal{T}-1]} g^{i}_{ref}$. Finally, the updated gradient $\tilde{g}$ computed using Eq. \ref{gradient} with $g_{ref}$ and $g$ can be used to update the model parameters.
After training the task $\mathcal{T}$, $D_{\mathcal{T}}^{ref}$ is added into the episodic memory $\mathcal{M}$. The training process will continue until the model is trained on all the tasks. 

Since the $l_2$-norms of the gradients $g$ and $g^{i}_{ref}$ are bounded, we can leverage a moments accountant to bound the privacy loss for a single task $\mathcal{T}$ as well as the privacy loss accumulated across all tasks. Let $\epsilon_{\mathcal{T}}$ be the privacy budget used to compute $g$ on $\mathcal{D}^{train}_{\mathcal{T}}$, and $\epsilon'_{i}$ is the privacy budget spent on computing the reference gradient $g^{i}_{ref}$ at each training task. The privacy budget used for a specific task $i \in [1, \mathcal{T})$, denoted as $\epsilon_{i}(\mathcal{T})$ and the total privacy budgets $\epsilon_{\mathcal{T}}^{all}$ of DP-AGEM accumulated until the task $\mathcal{T}$ can be computed in the following lemma.






\begin{lemma} Until the task $\mathcal{T}$, 1) the privacy budget used for a specific and previously learned task $i \in [1, \mathcal{T}]$ is: $\epsilon_{i}(\mathcal{T}) = \epsilon_{i} + (\mathcal{T} - i) \epsilon'_i$, and 2) the total privacy budget $\epsilon_{\mathcal{T}}^{all}$ of DP-AGEM is: $\epsilon_{\mathcal{T}}^{all} = \sum_{i = 1}^{\mathcal{T}} \epsilon_{i}(\mathcal{T})$.

\label{cdp}
\end{lemma}

\begin{proof} 
We use induction to prove Lemma \ref{cdp}.

When $\mathcal{T} = 1$, there is an empty episodic memory; therefore, $\epsilon_1(1) = \epsilon_1$ and $\epsilon_1^{all} = \epsilon_1 = \epsilon_1(1)$. 

Hence, Lemma \ref{cdp} is true for $\mathcal{T} = 1$.
Assuming that Lemma \ref{cdp} is true for $\mathcal{T} = k$, so we have $\epsilon_{i}(k) = \epsilon_{i} + (k - i) \epsilon'_i$ and  $\epsilon_{k}^{all} = \sum_{i = 1}^{k} \epsilon_{i}(k)$. We need to show that Lemma \ref{cdp} is true for $\mathcal{T} = k+1$.

We have: $\epsilon_{i}(k+1) = \epsilon_{i}(k) + \epsilon'_i = \epsilon_{i} + (k - i) \epsilon'_i + \epsilon'_i = \epsilon_{i} + (k+1- i) \epsilon'_i$,
and $\epsilon_{k+1}^{all} = \sum_{i = 1}^{k} \epsilon_{i}(k) + \epsilon_{k+1} + \sum_{i=1}^{k}  \epsilon'_i 
= \sum_{i = 1}^{k+1} \Big( \epsilon_i + (k+1-i) \epsilon'_i \Big)  = \sum_{i = 1}^{k+1} \epsilon_{i}(k+1)$. 
Thus, Lemma \ref{cdp} holds.  

\end{proof}

\textbf{Two Levels of DP Protection}. In Lemma \ref{cdp}, based on our definition of continual adjacent databases (Def. \ref{CAD}), it is essential that there are two levels of DP protection provided to an arbitrary data sample, as follows. Until the task $\mathcal{T} \in [1, N]$: \textbf{(1)} Given the DP budget $\epsilon_i(\mathcal{T})$ for a specific task $i \in [1, \mathcal{T}]$, the participation information of an arbitrary data sample in the task $i$ is protected under a $(\epsilon_i(\mathcal{T}), \delta)$-DP given the released parameters $\theta$. This can be presented as: $Pr[\text{DP-AGEM}(\mathcal{D}_i) = \theta] \leq e^{\epsilon_i(\mathcal{T})} Pr[\text{DP-AGEM}(\mathcal{D}'_i) = \theta] + \delta$, for any adjacent databases $\mathcal{D}_i$ and $\mathcal{D}'_i$; and \textbf{(2)} The participation information of an arbitrary data sample in the whole training data $(\mathcal{D} = \cup_{i = 1}^{\mathcal{T}} \mathcal{D}^{train}_i, \mathcal{M} = \cup_{i = 1}^{\mathcal{T}} \mathcal{D}^{ref}_i)$ is protected under a $(\epsilon_{\mathcal{T}}^{all}, \delta)$-DP given the released parameters $\theta$. This can be presented as:  $Pr[\text{DP-AGEM}(\mathcal{D}, \mathcal{M}) = \theta] \leq e^{\epsilon_\mathcal{T}^{all}} Pr[\text{DP-AGEM}(\mathcal{D}', \mathcal{M}') = \theta] + \delta$, for any continual adjacent databases $(\mathcal{D}, \mathcal{M})$ and $(\mathcal{D}', \mathcal{M}')$. This is fundamentally different from existing works \cite{farquhar2019differentially,phan2019private}, which do not provide any formal DP protection in CL.

Although DP-AGEM can preserve DP in CL, it suffers from a large privacy budget accumulation across tasks with an $O(\mathcal{T}^2)$ for $\epsilon^{all}_{\mathcal{T}}$. This is impractical in the real world with a loose DP protection. To address this, we present an algorithm to tighten the DP loss.

\textbf{DP-CL Algorithm}. Our DP-CL (Alg.~\ref{dp-agem-alg} and Figure~\ref{dp-agem}) takes a sequence of tasks $T = [t_i]_{i \in [1, N]}$ and dataset $\mathcal{D} = \cup_{i=1}^N \mathcal{D}_i$ as inputs. 
All samples in $D_{\mathcal{T}}^{train}$ are used to compute the proposed gradient update $g$ on task $\mathcal{T}$ with a sampling rate $p$ (Line 6). We clip  $g$ so that its $l_2$-norm is bounded by a predefined gradient clipping bound $\beta$. Then we add a random Gaussian noise $\mathcal{N}(0, \sigma^2 \beta^2 I)$ into $g$ with a predefined noise scale $\sigma$ (Line 9). Note that after training the task $\mathcal{T}$, samples in  $D_{\mathcal{T}}^{ref}$ are added to the episodic memory $\mathcal{M}$ as a mini-memory block $\mathcal{M}_{\mathcal{T}}$ (Lines 17, 24-26). 
To reduce the privacy budget accumulated over the number of tasks, we limit the access to seen data of previous tasks by using a randomly selected mini-memory block $\mathcal{M}_i$ ($i < \mathcal{T}$) from $\mathcal{M}$ to compute $g_{ref}$ (Lines 20-23).
We clip $g_{ref}$ by the gradient clipping bound $\beta$ and then add a random Gaussian noise $\mathcal{N}(0, \sigma^2 \beta^2 I)$ to $g_{ref}$ (Line 14). The updated gradient $\tilde{g}$ is computed by  Eq.~\ref{gradient} (Line 15).
Then $\tilde{g}$ is used to update the model parameters $\theta$ (Line 16).
The privacy budgets in our DP-CL algorithm can be bounded in the following lemma. 



\begin{algorithm}[t] 
\caption{DP in Continual Learning (DP-CL) Algorithm}\label{dp-agem-alg}
\begin{algorithmic}[1]
\STATE \textbf{Input:} Number of tasks $N$, dataset $\mathcal{D} = \cup_{i=1}^N  \mathcal{D}_i$, gradient clipping bound $\beta$, objective function $\mathcal{L}$
\STATE Initialize model $\theta$, episodic memory $\mathcal{M} = \emptyset$,  moments accountant $\mathbb{M}$
\FOR{$\mathcal{T} = \{1, ..., N\}$} 
\STATE $D_{\mathcal{T}}^{train} \sim \mathcal{D}_{\mathcal{T}}, D_{\mathcal{T}}^{ref} \sim \mathcal{D}_{\mathcal{T}}$ 
 s.t. $D_{\mathcal{T}}^{train} \cup D_{\mathcal{T}}^{ref} = \mathcal{D}_{\mathcal{T}}$, $D_{\mathcal{T}}^{train} \cap D^{ref}_{\mathcal{T}} = \emptyset$
 \FOR{each iteration $e=0, 1, 2, \ldots$ }
 \STATE $\mathcal{D}_{\mathcal{T}}^e \leftarrow$ Take random samples in $\mathcal{D}_{\mathcal{T}}^{train}$ with a sampling rate $p$  
\FOR{$(x, y) \in D^{e}_{\mathcal{T}}$}
            \STATE $g \leftarrow \nabla_{\theta} \mathcal{L} (f_{\theta} (x), y)$ \\
            \STATE $ g \leftarrow \text{ClipGrad}(g, \beta) +\mathcal{N}(0, \sigma^{2}\beta^{2}I) $
            \IF {$ \mathcal{T}=1$}
            \STATE $ \tilde{g} \leftarrow g$
            \ELSE  
            \STATE $ g_{ref} \leftarrow \text{CalGref}(\mathcal{M}, \mathcal{T}) $
        \STATE $ g_{ref} \leftarrow \text{ClipGrad}(g_{ref}, \beta) +\mathcal{N}(0, \sigma^{2}\beta^{2}I) $
        \STATE Compute $\tilde{g}$ with Eq.~\ref{gradient}
    \ENDIF
    \STATE $ \theta \leftarrow \theta - \alpha\tilde{g}$
\ENDFOR
\ENDFOR
\STATE $\mathcal{M} \leftarrow \textbf{UpdateEpsMem} (\mathcal{M}, D_{\mathcal{T}}^{ref}, \mathcal{T})$
\ENDFOR
\STATE print $\mathbb{M}$.\verb|get_priv_spent()|
\STATE \textbf{Output:} $(\epsilon, \delta)$-DP-CL $\theta$, $\mathbb{M}$

\STATE \textbf{\text{CalGref}}($\mathcal{M}, \mathcal{T}$):
\begin{ALC@g}
\STATE Randomly choose $\mathcal{M}_i$ from $\mathcal{M}$, where $i < \mathcal{T} $  
\STATE ($x^{ref}, y^{ref}) \sim \mathcal{M}_i $ 
    \STATE return  $g_{ref} = \nabla_{\theta} \mathcal{L} (f_{\theta} (x^{ref}), y^{ref})$
\end{ALC@g}


\STATE \textbf{\text{UpdateEpsMem}}($\mathcal{M}, D_{\mathcal{T}}^{ref}, \mathcal{T}$):
\begin{ALC@g}
\STATE $\mathcal{M}_{\mathcal{T}} \leftarrow D_{\mathcal{T}}^{ref}$
\STATE return  $ \mathcal{M} \cup \mathcal{M}_{\mathcal{T}}$
\end{ALC@g}

\STATE \textbf{\text{ClipGrad}}($g, \beta$):
\begin{ALC@g}
\STATE return $\pi(g, \beta) = g \cdot \min \Big(1, \frac{\beta}{ \| g \|} \Big) $
\end{ALC@g}
\end{algorithmic} 
\end{algorithm} 




\begin{lemma} Until the task $\mathcal{T}$, 1) the privacy budget used for a specific and previously learned task $i \in [1, \mathcal{T}]$ is: $\epsilon_{i}(\mathcal{T}) = \epsilon_{i} +  \epsilon'_i$, where $\epsilon'_i$ is the privacy budget used for a randomly chosen mini-memory block from $\mathcal{M}$ to compute $g_{ref}$ at task $i$, and 2) the total privacy budget $\epsilon_{\mathcal{T}}^{all}$ of DP-CL is: $\epsilon_{\mathcal{T}}^{all} = \sum_{i = 1}^{\mathcal{T}} \epsilon_{i}(\mathcal{T})$.
\label{lmrandom}
\end{lemma}



\begin{proof} 
We use induction to prove Lemma \ref{lmrandom}.
When $\mathcal{T} = 1$, there is an empty episodic memory; therefore, $\epsilon_1(1) = \epsilon_1$, $\epsilon_1^{all} = \epsilon_1 = \epsilon_1(1)$. Hence, Lemma \ref{lmrandom} is true for $\mathcal{T} = 1$.

Assuming that Lemma \ref{lmrandom} is true for $\mathcal{T} = k$, so we have $\epsilon_{i}(k) = \epsilon_{i} +  \epsilon'_i$ and  $\epsilon_{k}^{all} = \sum_{i = 1}^{k} \epsilon_{i}(k)$. We need to show that Lemma \ref{lmrandom} is true for $\mathcal{T} = k+1$.
We have: $\epsilon_{i}(k+1) =  \epsilon_{i} +  \epsilon'_i$,
and $\epsilon_{k+1}^{all} = \sum_{i = 1}^{k} \epsilon_{i}(k) + \epsilon_{k+1} +  \epsilon'_{k+1}  = \sum_{i = 1}^{k+1} \epsilon_{i}(k+1)$. 
Consequently, Lemma \ref{lmrandom} does hold. 

\end{proof} 

It is obvious that our DP-CL algorithm significantly reduces the privacy consumption to $O(\mathcal{T})$, which is linear to the number of training tasks. In addition, our sampling approach to compute $g_{ref}$ is unbiased, since the expectation for any data sample selected to compute $g_{ref}$ is the same: $\forall x \in \mathcal{M}, \mathbb{E}(x \in \mathcal{M}_i) = q/(\mathcal{T} - 1)$. In our experiment, we will show that DP-CL outperforms the baseline approach DP-AGEM.

\section{Experimental Results} 

\begin{figure*}[t]
\centering
\subfloat[\centering Privacy Accumulation]{\label{accumulation}\includegraphics[scale=0.33]{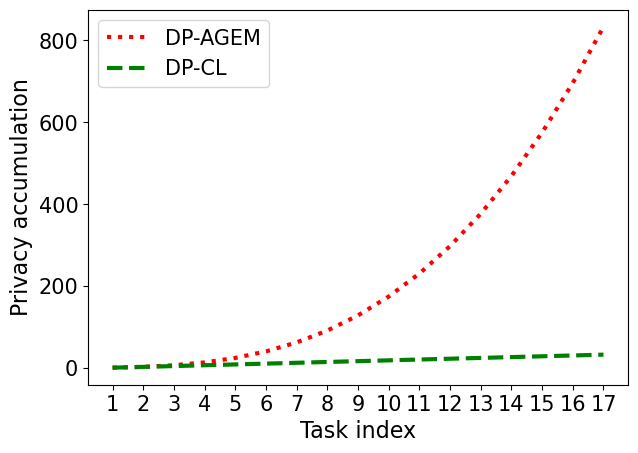}}
 \subfloat[Permuted MNIST Dataset]{\label{rd}\includegraphics[scale=0.33]{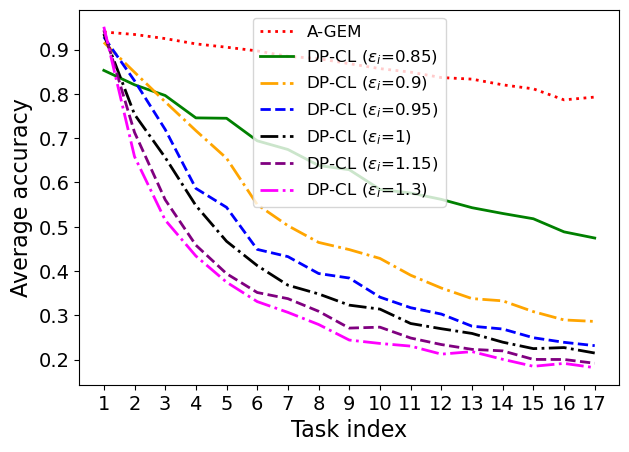}} 
\subfloat[Split CIFAR Dataset]{\label{cifar}\includegraphics[scale=0.33]{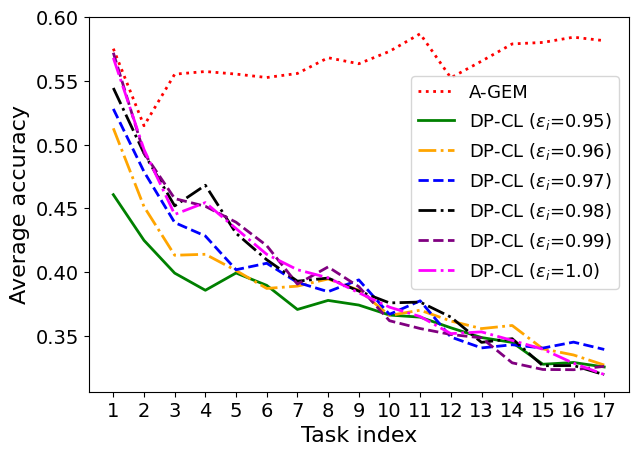}} 
\caption{Theoretical analysis for privacy accumulation (a); and Average accuracy over $17$ tasks of A-GEM and DP-CL algorithms with varying $\epsilon_i$ ($\epsilon'_i = 1.47$ and $\delta = 10^{-4}$ for all tasks) on the permuted MNIST dataset (b) and the permuted CIFAR-100 dataset (c).} 
\label{lm}
\end{figure*}

\begin{table*}[t] 
\caption{Forgetting measure (F), worst-case F, and LCA results for the MNIST dataset. The lower F, worst-case F, and the higher LCA the better.  } 
\label{lca}
\begin{center}
\begin{tabular}{|c|c|c|c|c|c|c|}
\hline
 \multicolumn{2}{|c|}{\multirow{2}*{\textbf{A-GEM} }} &   Forgetting (F) & Worst-case F & LCA  \\
\cline{3-5}   \multicolumn{2}{|c|}{}  &  $0.166 \pm 0.0070$ & $0.272 \pm 0.0086$ & $0.481 \pm 0.0051$   \\
\hline
 \multirow{6}*{\makecell{  \textbf{DP-CL} \\ ($\epsilon'_i = 1.47$ and $\delta = 10^{-4}$ for all tasks)  }  }  
& $\epsilon_i$ = 0.85 & $ 0.401 \pm$ 0.0070 & $ 0.586 \pm 0.0191$  & $ 0.146 \pm 0.0077$   \\
\cline{2-5}
 & $\epsilon_i$ = 0.9 &  $ 0.657\pm 0.0099$  & $ 0.809 \pm 0.0110$  & $ 0.123 \pm 0.0039$    \\
\cline{2-5}
& $\epsilon_i$ = 0.95 &  $ 0.713 \pm 0.0060$  & $ 0.840 \pm 0.0186 $  & $ 0.120 \pm 0.0038$  \\
\cline{2-5}
& $\epsilon_i$ = 1.0 &  $ 0.750\pm 0.0017$  & $ 0.851 \pm 0.0081$  & $0.119 \pm 0.0115$  \\
\cline{2-5}
& $\epsilon_i$ = 1.15 &  $ 0.782 \pm 0.0017$  & $ 0.863 \pm 0.0061$  & $ 0.124\pm 0.0013$   \\
\cline{2-5}
 & $\epsilon_i$ = 1.30 &  $ 0.796 \pm 0.0023$  & $ 0.864\pm 0.0077$  & $ 0.121 \pm 0.0021$ \\
\hline

\end{tabular} 
\end{center} 
\end{table*}

\begin{table*}[t] 
\caption{Forgetting measure (F), worst-case F, and LCA results for the Split CIFAR dataset. The lower F, worst-case F, and the higher LCA the better.  } 
\label{lca-cifar}
\begin{center}
\begin{tabular}{|c|c|c|c|c|c|c|}
\hline
 \multicolumn{2}{|c|}{\multirow{2}*{\textbf{A-GEM} }} &   Forgetting (F) & Worst-case F & LCA  \\
\cline{3-5}   \multicolumn{2}{|c|}{}  &  $0.089 \pm 0.0163$ & $0.188\pm 0.0317$ & $0.348 \pm 0.0111$   \\
\hline
 \multirow{6}*{\makecell{  \textbf{DP-CL} \\ ($\epsilon'_i = 1.47$ and $\delta = 10^{-4}$ for all tasks)  }  }  
& $\epsilon_i$ = 0.95 & $ 0.149 \pm 0.0123$ & $ 0.314 \pm 0.0057$  & $0.262 \pm 0.0058$   \\
\cline{2-5}
 & $\epsilon_i$ = 0.96 &  $ 0.181\pm 0.0193$  & $ 0.335 \pm 0.0421$  & $ 0.259 \pm 0.0130$    \\
\cline{2-5}
& $\epsilon_i$ = 0.97 &  $ 0.196 \pm 0.0194$  & $ 0.377 \pm 0.0174 $  & $ 0.266 \pm 0.0111$  \\
\cline{2-5}
& $\epsilon_i$ = 0.98 &  $ 0.239\pm 0.0162$  & $ 0.428 \pm 0.0701$  & $0.266 \pm 0.0008$  \\
\cline{2-5}
& $\epsilon_i$ = 0.99 &  $ 0.249 \pm 0.0097$  & $ 0.435 \pm 0.0432$  & $ 0.259\pm 0.0053$   \\
\cline{2-5}
 & $\epsilon_i$ = 1.0 &  $ 0.262 \pm 0.031$  & $ 0.455\pm 0.0452$  & $ 0.263 \pm 0.0096$ \\
\hline

\end{tabular} 
\end{center}
\end{table*}

We have conducted  experiments on the permuted MNIST dataset \cite{goodfellow2013empirical} and the Split CIFAR dataset \cite{zenke2017continual}. The permuted MNIST dataset is a variant of the MNIST dataset \cite{lecun1998mnist} of handwritten digits. The permuted MNIST dataset has $55,000$ images involving ten-digit classification, where each task consists of different random permutations of the input pixels in the images. The Split CIFAR \cite{zenke2017continual}  is a split version of the original
CIFAR-100 dataset \cite{krizhevsky2009learning}.  There are $20$ disjoint subsets, where each subset is constructed by randomly sampling $5$ classes without replacement from a total of $100$ classes. Our validation focuses on shedding light on the interplay between model utility and privacy loss of preserving DP in CL. Our code and datasets are available on Github\footnote{\url{https://github.com/PhungLai728/DP-CL}}.

\textbf{Baseline Approaches}. We evaluate our DP-CL algorithm and compare it with A-GEM \cite{chaudhry2018efficient}, one of the state-of-the-art CL algorithms. Note that A-GEM does not preserve DP; therefore, we only use A-GEM to show the upper-bound model performance. We apply four well-known metrics, including the \textit{average accuracy}, the \textit{average forgetting measure} \cite{chaudhry2018riemannian}, the \textit{worst-case forgetting measure} \cite{chaudhry2018efficient}, and the \textit{learning curve area} (LCA) \cite{chaudhry2018efficient}, to evaluate our mechanism. 


\textbf{Model Configuration}.
In the permuted MNIST dataset, we use a fully connected network with two hidden layers of $256$ hidden neurons. Given a stream of $17$ tasks, the model is optimized via stochastic gradient descent with a learning rate $\alpha = 0.1$. In computing $g_{ref}$, the batch-size is set to $100$ for each training task and $50$ for the mini-memory block. The number of runs for each experiment is $3$. The noise scale $\sigma = 1$ and the gradient clipping bound $\beta = 0.1$. In the Split CIFAR dataset, a reduced ResNet-18 \cite{lopez2017gradient,he2016deep} with 3 times less feature maps across all the layers. The network has a final linear classifier for prediction in the  Split CIFAR dataset. The batch-size is set to $10$ in each training task. Other hyperparameters, e.g., learning rate, noise scale, gradient clipping bound, etc., are the same as in the permuted MNIST dataset experiment. The number of runs for each experiment is $5$.

$\bullet$ \textbf{Comparing Privacy Accumulation}.
Since the number of data samples and the sampling rate remain the same for every task, the privacy budgets $\epsilon_i$ and $\epsilon'_{i}$ can be the same for every task. Therefore, for the sake of clarity without loss of generality, in this privacy accumulation comparison between DP-AGEM and our DP-CL algorithm, we draw different random Gaussian values with (mean $=1$, std $=0.02$) and assign the generated values as the privacy budget $\epsilon_i$ and $\epsilon'_{i}$ for $17$ tasks. 

Figure \ref{accumulation} illustrates how privacy loss accumulates over $17$ tasks in DP-AGEM and our DP-CL algorithm. Our algorithm achieves a notably tighter privacy budget compared with DP-AGEM, which accesses data samples from the whole episodic memory to compute $g_{ref}$. When the number of tasks increases, DP-AGEM's privacy budget exponentially increases. In contrast, our approach's privacy budget slightly increases and is linear to the number of tasks or training steps.

$\bullet$ \textbf{Privacy Loss and Model Utility}. From our theoretical analysis, DP-AGEM suffers from a huge privacy budget accumulation over tasks. Therefore, we only compare our DP-CL algorithm and the noiseless A-GEM model for the sake of simplicity. 

As shown in Figure \ref{rd} and  \ref{cifar}, our proposed method achieves a comparable average accuracy with the noiseless A-GEM model at the first task. In the permuted MNIST dataset, when the number of tasks increases, the average accuracy of our DP-CL drops faster than the average accuracy of the A-GEM model. For example, at task $17$-th, A-GEM's average accuracy  drops to $79.3\%$, while DP-CL's average accuracy drops to $47.5\%$ with a tight privacy budget $\epsilon_i = 0.85$. 
When the privacy budget increases, the average accuracy gap between our model and the noiseless A-GEM is larger, indicating that preserving DP in CL may increase the catastrophic forgetting. This phenomenon is further clarified by the measures of forgetting, worst-case forgetting, and LCA (Table~\ref{lca}). At $\epsilon_i = 0.85$, forgetting, worst-case forgetting, and LCA are $0.401$, $0.586$, and $0.146$ respectively in DP-CL. After that, the forgetting and worst-case forgetting  significantly increase, and LCA moderately decreases in DP-CL. 

In the Split CIFAR dataset, when the number of tasks increases, the average accuracy of DP-CL drops quickly while the average accuracy of the A-GEM model fluctuates. For instances, A-GEM's average accuracy is $57.5\%$ at the first task, drops to $51.5\%$ at the second task, and is $58.1\%$ at the last task. Meanwhile, DP-CL's average accuracy is $56.8\%$ at the first task, and gradually drops to $31.9\%$ at the last task with a tight privacy budget $\epsilon_i = 1.0$. The fluctuation phenomenon in the A-GEM model is probably due to the curse of dimension in which there are $2,500$ training examples, which is much smaller than the number of trainable parameters in the ResNet-18, i.e., $11$ million.     
Different from the permuted MNIST dataset, in the Split CIFAR dataset, when the privacy budget increases, the average accuracy gap between DP-CL and the noiseless A-GEM is smaller, especially at the first task. For instance, at the first task, the gap are $11.4\%$, $6.3\%$, $4.7\%$, $3.1\%$, $0.3\%$, and $0.7\%$ when the values of $\epsilon_i$ are $0.95$, $0.96$, $0.97$, $0.98$, $0.99$, and $1.0$, respectively. This shows the trade-off between privacy budget and model utility in which when we spend more privacy budget, the model accuracy improves. The gap between  DP-CL's average accuracy and A-GEM's average accuracy are significantly bigger when the number of tasks increases, but the difference among different privacy budgets decreases. For instance, at the last task, the gap are $25.6\%$, $25.4\%$, $24.2\%$, $26.2\%$, $25.5\%$, and $26.2\%$ when the values of $\epsilon_i$ are $0.95$, $0.96$, $0.97$, $0.98$, $0.99$, and $1.0$, respectively.  As shown in Table \ref{lca-cifar}, when the privacy budget increases, the forgetting and worst-case forgetting significantly increase, while the LCA slightly fluctuates around $[0.259,0.266]$. This further confirms our observations in the MNIST dataset in which preserving DP in CL may increase the catastrophic forgetting. 

\textbf{Key observations}. From our preliminary experiments, we obtain the following observations. \textbf{(1)} Merely incorporating the moments accountant into A-GEM causes a large privacy budget accumulation. \textbf{(2)} Although our DP-CL algorithm can preserve DP in CL, optimizing the trade-off between model utility and privacy loss in CL is an open problem since the privacy noise can worsen the catastrophic forgetting.


\section{Conclusion and Future Work}
In this paper, we established the first formal connection between DP and CL. We combine the moments accountant and A-GEM in a holistic approach to preserve DP in CL in a tightly accumulated privacy budget. Our model shows promising results under strong DP guarantees in CL and opens a new research line to optimize the model utility and privacy loss trade-off. One of the immediate questions is how to align the privacy noise with the catastrophic forgetting under the same privacy protection. We also plan to examine our approach to a broader range of models and datasets, especially under adversarial attacks \cite{shokri2017membership,carlini2020extracting}, and heterogeneous and adaptive privacy-preserving mechanisms \cite{10.5555/3367471.3367704,phan2017adaptive,10.1145/3447548.3467268}. Our work further highlights an open direction of quantifying the privacy risk given a diverse correlation among tasks. Learning a highly related task can further disclose the private information in another task, and vice-versa. \vspace{-10pt}

\section*{Acknowledgments} \vspace{-10pt}
The authors gratefully acknowledge the support from the National Science Foundation (NSF) grants NSF CNS-1935928/1935923, CNS-1850094, IIS-2041096/2041065. \vspace{-10pt}

\bibliographystyle{splncs04}
\bibliography{DP2LM}

\end{document}